\documentclass[a4paper]{article}
\usepackage[spanish]{babel} % Soporte para idioma español y traducciones
\usepackage{amsmath}
\usepackage{amssymb}
\usepackage[ruled,vlined,spanish]{algorithm2e} % Sirve para incluir algoritmos y pseudocódigos en el documento. Los parámetros "ruled" y "vlined" establecen el estilo de presentación del algoritmo, mientras que "spanish" indica que se utilice el idioma español para las etiquetas y formatos predeterminados.
\usepackage{graphicx}
\usepackage[hidelinks]{hyperref}
\usepackage{subcaption} % Para subfiguras dentro de una figura
\usepackage[utf8]{inputenc}
\usepackage[T1]{fontenc}
\usepackage{appendix}

\usepackage{booktabs} % Para mejorar el formato de las tablas (por ejemplo, \toprule, \midrule, \bottomrule)
\usepackage{caption} % Para mejorar la gestión del caption y su personalización
\addto\captionsspanish{} % Cambia la palabra "Cuadro" por "tabla" en el caption de las mismas
\usepackage{listings}
\usepackage{xcolor}

\usepackage{geometry}
\title{\LARGE Underdamped Particle Swarm Optimization \\
\Large Python Implementation}
\author{Hernández Rodríguez, Matías Ezequiel}
\date{March 2025}

\begin{document}

\maketitle

\begin{abstract}
This article presents Underdamped Particle Swarm Optimization (UEPS), a novel metaheuristic inspired by both the Particle Swarm Optimization (PSO) algorithm and the dynamic behavior of an underdamped system. The underdamped motion acts as an intermediate solution between undamped systems, which oscillate indefinitely, and overdamped systems, which stabilize without oscillation. In the context of optimization, this type of motion allows particles to explore the search space dynamically, alternating between exploration and exploitation, with the ability to overshoot the optimal solution to explore new regions and avoid getting trapped in local optima.

First, we review the concept of damped vibrations, an essential physical principle that describes how a system oscillates while losing energy over time, behaving in an underdamped, overdamped, or critically damped manner. This understanding forms the foundation for applying these concepts to optimization, ensuring a balanced management of exploration and exploitation. Furthermore, the classical PSO algorithm is discussed, highlighting its fundamental features and limitations, providing the necessary context to understand how the underdamped behavior improves PSO performance.

The proposed metaheuristic is evaluated using benchmark functions and classic engineering problems, demonstrating its high robustness and efficiency.
\end{abstract}

\section{Introduction}

Optimization is a fundamental branch of science that seeks to find the best possible solutions to complex problems through the use of various methods and algorithms. In this context, there are multiple approaches that can be classified into optimization algorithms, iterative methods, and heuristics. Optimization algorithms are known for their ability to reach a solution in a finite number of steps, such as the famous \textit{Simplex algorithm} for linear programming or the variants of combinatorial and quadratic methods. On the other hand, \textit{iterative methods} are common in nonlinear programming problems, where optimization is achieved by repeating steps that gradually improve the solution. However, in many cases, classical methods may be insufficient due to the high complexity of the problems, making it necessary to resort to \textit{heuristics}. These techniques, such as \textit{genetic algorithms} or \textit{particle swarm optimization}, are particularly useful for addressing problems where the search for the exact solution is computationally expensive or even unfeasible, providing approximate solutions with relatively low computational effort \cite{Ha}.

Among these heuristic optimization techniques, PSO has proven to be an effective and flexible approach. This algorithm is inspired by the collective behavior of particle swarms, where particles interact with each other and their environment to find optimal solutions in a search space. However, the traditional PSO algorithm can present limitations when facing complex problems, particularly regarding the ability to effectively explore the entire search space, and the tendency of particles to get trapped in local optima \cite{Ken}.

In this article, we present a variation of PSO, called \textit{Underdamped Particle Swarm Optimization (UEPS)}. This algorithm incorporates the concept of \textit{underdamped motion}, which describes a dynamic system where particles do not stabilize immediately but experience oscillations before reaching equilibrium. This type of motion is an intermediate solution between the behavior of an undamped system (which oscillates indefinitely) and an overdamped system (which stabilizes without oscillations). In the context of optimization, UEPS allows particles to explore the search space more dynamically, alternating between \textit{exploration} and \textit{exploitation}. The ability to "overshoot" the optimal solution during oscillations enables particles to escape local optima and explore new regions of the space, improving the algorithm's robustness.

Section 2 provides a detailed overview of the fundamentals of \textit{mathematical optimization}. Before addressing the mathematical formulation of the proposed metaheuristic, it is essential to understand the underlying physical principles that inspire this approach. For this reason, Section 3 offers an introductory analysis of the concept of \textit{damped vibrations}, a fundamental topic in physics that describes how a system oscillates while gradually losing energy over time. This behavior is classified into three regimes: underdamped, overdamped, and critically damped. By understanding how an underdamped system can oscillate and explore different states before stabilizing, we can apply these principles to the design of optimization algorithms, allowing for efficient management of \textit{exploration} and \textit{exploitation} of the search space. This connection between a physical concept and its mathematical application provides not only a solid theoretical foundation but also a meaningful pedagogical approach, as it facilitates understanding how abstract physical concepts can be used to solve complex optimization problems.

Section 4 presents a review of the classical PSO algorithm, with an analysis of its fundamental features and inherent limitations. This review is crucial for understanding how the incorporation of underdamped behaviors in the proposed variant, the UEPS algorithm, can significantly improve performance and the algorithm's exploration capacity, more effectively addressing complex optimization problems.

Sections 5 and 6 are dedicated to the mathematical formulation of the UEPS algorithm and its respective \textit{pseudocode}. These sections provide the necessary details to understand how the principles of damped vibrations are implemented within the PSO framework, as well as present the algorithm's structure that manages the particle dynamics during the search process.

Section 7 is dedicated to the numerical results obtained through the validation of the proposed metaheuristic. In this section, the performance of UEPS is validated using benchmark functions, both for optimization problems with and without constraints, such as the \textit{Ackley} and \textit{Rosenbrock} functions, among others. Additionally, a practical application of the algorithm to the \textit{design problem of a pressure vessel} is presented, demonstrating how the underdamped approach outperforms other known metaheuristics.

Section 8 presents the conclusions drawn from the results obtained and future work perspectives. The study of new applications and improvements to the metaheuristic opens up a range of possibilities for its expansion into various areas of optimization. Additionally, an \textit{appendix} is included that explains in detail the implementation of UEPS in Python, providing the reader with the opportunity to replicate the experimental results presented, as well as the flexibility to tackle other optimization problems that may arise in future studies.

Finally, we provide the Python implementation of the UEPS algorithm in the appendix of the document.

\section{Mathematical Optimization}

In this section, we provide a brief overview of the mathematical formulation of the concept of optimization. Optimization is presented as a fundamental tool in various fields of science and engineering, with the goal of finding the best possible solution within a given set of constraints and objectives. We will focus on the minimization of an objective function subject to equality and inequality constraints, a problem that is formulated in the following mathematical statement:

\begin{equation} \label{eq1}
\begin{aligned}
&\ \min_{\mathbf{x} \in \mathbb{R}^n} f(\mathbf{x}) \\
&\ \text{subject to:} \\
&\ h_j(\mathbf{x}) = 0, \quad j = 1, \dots, M, \\
&\ g_k(\mathbf{x}) \leq 0, \quad k = 1, \dots, N,
\end{aligned}
\end{equation}
where $f,h_j,g_k:\mathbb{R}^n\longrightarrow\mathbb{R}$\footnote{Maximizing a functional, $f$, is equivalent to minimizing its negative, i.e., $-f$; therefore, all methods applicable to (\ref{eq1}) are equally valid for maximization problems.}. In this scenario, the variables $x_i$ ($i=1,\dots,n$) become our decision agents, manipulating an \textit{objective function} or \textit{cost function} $f$, while the \textit{equality and inequality constraints}, represented by $h_j$ and $g_k$, define the \textit{search space} $\Omega \subseteq \mathbb{R}^n$ and the set of solutions $f(\Omega)$. The element $\mathbf{x}^* \in \Omega$ that minimizes (or maximizes) $f$ is called the \textit{optimum}.

There are different approaches to addressing optimization problems with constraints using metaheuristics \cite{Coello}. One strategy, the most intuitive approach, involves transforming the problem (\ref{eq1}) into the following problem:

\begin{align} \label{eqOpt1}
    \min_{\mathbf{x}\in\mathbb{R}^n} \quad  F(\mathbf{x})   
\end{align}
where $F(\mathbf{x})=\displaystyle\sum_{i=1}^N r_i \max\{g_i(\mathbf{x}), 0\} + \displaystyle\sum_{j=1}^P c_j |h_j(\mathbf{x})|,$ the functions $\max\{g_i(\mathbf{x}),0\}$ and $|h_j(\mathbf{x})|$ measure the degree to which the constraints are violated, and $r_i$ and $c_j$ are known as penalty parameters.

In the other strategy,

\begin{equation}\label{eqOpt2}
F(\mathbf{x}) =
\begin{cases}
f(\mathbf{x}) & \text{if the solution is feasible;} \\
K \left( 1 - \frac{s}{m} \right) & \text{otherwise,}
\end{cases}
\end{equation}
where $s$ is the number of satisfied constraints, $m$ is the total number of constraints (equalities and inequalities), and $K$ is a large constant ($K = 1\times 10^9$) \cite{Mora}.

\section{Damped Vibrations}

A \textit{damped system} is a dynamic system in which the energy of vibrations is dissipated over time due to resistance (damping). This concept is essential to understand how vibrations are managed in various physical systems. The differential equation that describes the motion of a damped system is as follows:

$$
m \cdot \ddot{x}(t) + c \cdot \dot{x}(t) + k \cdot x(t) = 0,
$$
where:
\begin{itemize}
	\item $ m $ is the mass of the system,
	\item $ c $ is the damping coefficient (resistance to vibrations),
	\item $ k $ is the spring constant (stiffness of the system),
	\item $ x(t) $ is the position of the system at time $ t $.
\end{itemize}

The solution to this equation depends on the relationship between the parameters $ m $, $ c $, and $ k $. In particular, the \textit{angular frequency} $ \omega_0 $ and the \textit{damping coefficient} $ \gamma $ are two key parameters that determine the behavior of the system.

\begin{enumerate}
	\item Angular frequency ($ \omega_0 $):
	   The \textit{angular frequency} is the frequency at which the system would oscillate if there were no damping (if $ c = 0 $). It is defined as:

	   $$
	   \omega_0 = \sqrt{\frac{k}{m}}.
	   $$

	   That is, it depends on the spring constant $ k $ and the mass $ m $. This frequency determines the rate of the natural oscillations of the system in the absence of damping.

	\item Damping coefficient $( \gamma $):
	   The \textit{damping coefficient} is a measure of the resistance the system presents to vibrations. In this case, it is defined as:

	   $$
	   \gamma = \frac{c}{2m},
	   $$
which means it depends on the damping coefficient $ c $ and the mass $ m $. The value of $ \gamma $ determines how quickly the system loses energy (how much the oscillations are "damped").
\end{enumerate}

\subsection{Types of Behavior According to $ \gamma $ and $ \omega_0 $}

\subsubsection{Overdamped}
The system is \textit{overdamped} when the damping coefficient $ c $ is so large that it prevents any oscillation. In this case, the roots of the characteristic equation are real and distinct, and the system returns to its equilibrium position monotonically.

\begin{itemize}
	\item Condition: $ \gamma > \omega_0 $
	\item Solution:
	  $$
	  x(t) = A \cdot e^{\lambda_1 t} + B \cdot e^{\lambda_2 t},
	  $$
	  where $ \lambda_1 $ and $ \lambda_2 $ are the real, distinct negative roots, and $ A $ and $ B $ are constants determined by the initial conditions.
\end{itemize}

\subsubsection{Critically Damped}
The system is \textit{critically damped} when the damping coefficient is exactly what is needed to avoid oscillations, and the system returns to equilibrium as quickly as possible without oscillating. This is the critical case, where the system does not oscillate but stabilizes quickly.

\begin{itemize}
	\item Condition: $ \gamma = \omega_0 $
	\item Solution:
	  $$
	  x(t) = (A + Bt) \cdot e^{-\gamma t},
	  $$
	  where $ A $ and $ B $ are constants determined by the initial conditions.
\end{itemize}

\subsubsection{Underdamped}
The system is \textit{underdamped} when the damping coefficient is small enough to allow the system to oscillate around the equilibrium position before stopping due to energy dissipation. This is the case we are interested in, as it has oscillations that decrease over time.

\begin{itemize}
	\item Condition: $ \gamma < \omega_0 $
	\item Solution:
	  $$
	  x(t) = A \cdot e^{-\gamma t} \cdot \cos(\omega_d t + \phi),
	  $$
	  where $ \omega_d = \sqrt{\omega_0^2 - \gamma^2} $ is the frequency of the damped oscillations, and $ \phi $ is a phase determined by the initial conditions.
\end{itemize}

\begin{figure}[ht]
\centering
\includegraphics[width=0.8\textwidth]{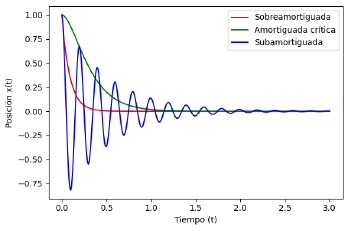}  % Adjust the image path and size
\caption{Types of damped vibrations.}  % Caption describing the image
\label{fig:vibraciones}  % Label for referencing the figure in the text
\end{figure}

Figure~\ref{fig:vibraciones} shows the dynamics of the three types of damped vibrations described above using $m = 0.1,$ $k = 100$, $A = 1,$ $B = 0$ and $\gamma = 10, 6$ and $2$ for overdamped, critically damped, and underdamped vibrations, respectively.

\section{Review of Classical PSO}

In this section, we will review PSO, which was proposed by \textit{Kennedy} and \textit{Eberhart} in 1995 \cite{Ken}. This algorithm is based on the collective behavior of a set of particles that move through the search space in search of the best solution. Each particle has a position and a velocity, and it seeks the best solution by combining its own knowledge with that of its neighbors.

\subsection{Particle Model}

In classical PSO, each particle $i$ in the swarm is described by the following properties:
\begin{itemize}
    \item The position of the particle $\mathbf{x}_i = (x_{i1}, x_{i2}, \dots, x_{id})$, where $d$ is the number of dimensions of the search space.
    \item The velocity of the particle $\mathbf{v}_i = (v_{i1}, v_{i2}, \dots, v_{id})$, which indicates the direction and magnitude of the particle's movement.
    \item The best known position of the particle, $\mathbf{p}_i = (p_{i1}, p_{i2}, \dots, p_{id})$, which is the best solution found by the particle during its trajectory.
    \item The global best position, $\mathbf{g} = (g_1, g_2, \dots, g_d)$, which is the best solution known by the entire swarm.
\end{itemize}

\subsection{Evolution of the Particles}

At each iteration of the algorithm, the position and velocity of each particle are updated according to the following equations:

\begin{equation}
    \mathbf{v}_i^{k+1} = w^t \cdot \mathbf{v}_i^k + c_1 \cdot \mathbf{r}_1 \cdot (\mathbf{p}_i - \mathbf{x}_i^k) + c_2 \cdot \mathbf{r}_2 \cdot (\mathbf{g} - \mathbf{x}_i^k)
\end{equation}

\begin{equation}
    \mathbf{x}_i^{k+1} = \mathbf{x}_i^k + \mathbf{v}_i^{k+1}
\end{equation}

Where:
\begin{itemize}
    \item $\mathbf{v}_i^k$ is the velocity of particle $i$ in iteration $k$. Empirical studies have shown that it is most effective to initialize the velocities to zero \cite{Enge}.
    \item $\mathbf{x}_i^k$ is the position of particle $i$ in iteration $k$.
    \item $w^t$ is the inertia factor that controls the rate of change of the particle's velocity. To balance exploration and exploitation, $w^t$ is dynamically adjusted as follows:

    $$w^k = w_{max} - (w_{max} - w_{min}) \frac{k}{t_{max}},$$
    where $w_{max}$ and $w_{min}$ are the maximum and minimum values of the inertia factor, and $t_{max}$ is the maximum number of iterations. Typical values that produce good results are $w_{max} = 0.9$ and $w_{min} = 0.4$ \cite{Shi, Zha}.
    
    \item $c_1$ and $c_2$ are the acceleration coefficients that determine the influence of the best local and global positions. These values are recommended to be in the interval $[1.8, 2]$ \cite{Shi}.
    \item $\mathbf{r}_1$ and $\mathbf{r}_2$ are vectors with randomly distributed numbers in the interval $[0,1]$.
\end{itemize}

The algorithm will iterate until a convergence criterion is met or the maximum number of iterations is reached.

\subsection{Features of PSO}

PSO has several important features that make it suitable for optimizing complex problems:
\begin{itemize}
    \item \textbf{Fast convergence:} PSO converges quickly to an optimal solution due to the interaction between the particles and the swarm.
    \item \textbf{Balanced exploration and exploitation:} The algorithm combines exploration of the search space with exploitation of solutions close to the best found.
    \item \textbf{Simplicity:} PSO is relatively easy to implement, making it an attractive option for high-dimensional optimization problems.
\end{itemize}

\section{Mathematical Formulation of UEPS}

The algorithm we propose is based on a position update similar to that of the particles in PSO, but modified to include a damped term that allows oscillation and exploration beyond local solutions.

\subsection{Particle Update}

The position update of each particle is performed using the following formula:

\begin{equation}
\mathbf{x}_i(t+1) = \mathbf{x}_i(t) + \mathbf{v}_i(t+1),
\end{equation}
where the velocity \( \mathbf{v}_i(t+1) \) is calculated as follows:

\begin{equation}
\mathbf{v}_i(t+1) = w^t \cdot \mathbf{v}_i(t) + \left[ A \cdot (2 - \cos(2\pi \cdot \mathbf{r_1})) \cdot \exp(-b \cdot t) \right] \cdot (\mathbf{x}^* - \mathbf{x}_i(t)) + \alpha \cdot (\mathbf{r_2} - 0.5),
\end{equation}
where

\begin{itemize}
    \item  \( A \): The oscillation amplitude, controls how far the particle can move in each iteration.
    \item  \( b \): The damping factor, controls the rate at which oscillations decrease.
    \item  \( \mathbf{r_1} \): A random vector with values between 0 and 1 used to generate oscillations.
    \item  \( \mathbf{x}^* \): The best global position known by the swarm.
    \item  \( \alpha \): A parameter that introduces an additional random disturbance to improve exploration.
    \item  \( \mathbf{r_2} \): A random vector between 0 and 1, used for the random disturbance.
    \item  \( 0.5 \): A vector with components equal to 0.5.
    \item  \( w^t \): The inertia factor as in PSO.
\end{itemize}

\section{Pseudocode of the Damped PSO Algorithm}

The pseudocode~\ref{alg:PSOSA} describes the UEPS algorithm, which, as we saw earlier, incorporates a damped oscillatory term to help convergence in complex search spaces.

\begin{algorithm}[H]
\caption{Pseudocode of the Damped PSO Algorithm}
\label{alg:PSOSA}
\SetKwInput{KwInput}{Input} 
\SetKwInput{KwOutput}{Output}

\KwInput{Objective function \( f(\mathbf{x}) \), lower bound \( L_b \), upper bound \( U_b \), parameters \( A \), \( b \), \( w_{\text{min}} \), \( w_{\text{max}} \), \( \alpha \), number of particles \( n \), number of iterations \( T \)}
\KwOutput{Best solution found \( \mathbf{x}_{\text{global best}} \) and its objective value \( f_{\text{global best}} \)}

\SetKwFunction{GlobalSearch}{Initialization}
\SetKwFunction{UpdatePosition}{UpdatePosition}
\SetKwFunction{EvaluateFitness}{EvaluateFitness}
\SetKwFunction{UpdateBest}{UpdateBest}
\SetKwFunction{UpdateGlobalBest}{UpdateGlobalBest}

\GlobalSearch{Initialize random positions \( \mathbf{X_i} \) and velocities \( \mathbf{V_i} \)}\;
\GlobalSearch{Calculate the fitness \( f(\mathbf{X_i}) \) for each particle}\;
\GlobalSearch{Set the best local and global positions}\;

\While{stopping criterion not met}{
    \For{each particle \( \mathbf{X_i} \)}{
        \UpdatePosition{Update the particle position using the velocity \( V_i \)}\;
        \EvaluateFitness{Evaluate the objective function at \( \mathbf{X_i} \)}\;
        \UpdateBest{Update the personal best position \( \mathbf{X_{\text{best}}} \) if necessary}\;
    }
    \UpdateGlobalBest{Update the global best position \( \mathbf{x_{\text{global best}}} \)}\;
}

\Return{$\mathbf{x_{\text{global best}}}$, $f_{\text{global best}}$}\;

\end{algorithm}

\section{Implementation and Numerical Results}
Based on the pseudocode~\ref{alg:PSOSA}, the implementation of the UEPS algorithm is straightforward and simple. To validate the effectiveness of our metaheuristic, we have developed a project in Python consisting of two main modules: \texttt{test\_functions.py} and \texttt{oeps.py}. The first of these modules includes various testing functions, such as the Rosenbrock function (both with and without constraints), as well as other classic optimization problems in engineering. The second module, which imports the first one, is responsible for the implementation of the UEPS metaheuristic, allowing the user to specify the optimization problem they wish to address.

This project has been designed to be flexible, allowing for the addition of new testing functions or optimization problems as needed. The source code of both modules is presented in the appendix of this paper.

Next, we will validate the proposed metaheuristic using standard test functions and two classic engineering optimization problems. For the test functions, we will compare the results obtained by our algorithm with the known solutions, as the optimal solutions have already been previously established. For the classic engineering problems, the results obtained will be compared with those from other well-known metaheuristics in the literature. The parameters used in each of the applications we will address are those specified in the implementation of the algorithm, detailed in the code included in the appendix of this paper.

\subsection{Unconstrained Problems}

\begin{figure}[ht]
\centering
\includegraphics[width=0.8\textwidth]{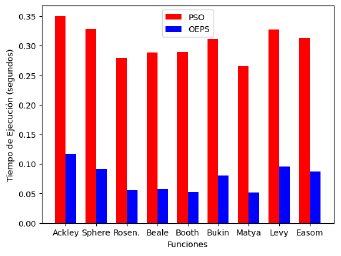}  % Ajusta la ruta y el tamaño de la imagen
\caption{Comparison of execution time between UEPS and PSO.}  % Aquí va el texto que describes en el caption
\label{fig:PSOvsOEPS}  % Etiqueta para referenciar la figura en el texto
\end{figure}

In this subsection, we will evaluate the performance of UEPS and PSO in optimizing unconstrained test functions. For all the problems considered, 10 runs were performed, using 50 particles and 100 iterations in each case. The average results obtained, using 6 significant digits, from minimizing eight test functions are presented in tables~\ref{tab:Ej_0} and \ref{tab:Ej_0_1} for UEPS and PSO, respectively. The approximations of both algorithms are very similar, except for the \textit{Beale} and \textit{Easom} functions, where UEPS was considerably superior to PSO. Figure~\ref{fig:PSOvsOEPS} shows the execution time demanded by each algorithm, demonstrating that UEPS performs better.

\begin{table}[ht]
\centering
\scriptsize
\begin{tabular*}{\textwidth}{l @{\extracolsep{\fill}}llll}
\toprule
Function & Formula & Global minimum & Range & Approximation \\
\midrule
Ackley & $-20 e^{-0.2\sqrt{0.5(x^2 + y^2)}} - e^{0.5(\cos 2\pi x + \cos 2\pi y)} + e + 20$ & (0,0) & $[-5,5]^2$ & $(0.000000, 0.000000)$ \\
\midrule
Sphere & $x^2 + y^2$ & (0,0) & $[-100,100]^2$ & $(0.000000, 0.000000)$ \\
\midrule
Rosenbrock & $(1-x)^2+100(y-x^2)^2$ & (1,1) & $[-10,10]^2$ & $(0.999997, 0.999995)$ \\
\midrule
Beale & $(1.5 - x + xy)^2 + (2.25 - x + xy^2)^2 + (2.625 - x + xy^3)^2$ & (3,0.5) & $[-4.5,4.5]^2$ & $(3.000000, 0.500000)$ \\
\midrule
Booth & $(1.5 - x + xy)^2 + (2.25 - x + xy^2)^2$ & (1,3) & $[-10,10]^2$ & $(1.000000, 3.000000)$ \\
\midrule
Matyas & $0.26(x^2 + y^2) - 0.48xy$ & (0,0) & $[-10,10]^2$ & $(0.000000, 0.000000)$ \\
\midrule
Levy & $\sin^2 3\pi x + (x - 1)^2 (1 + \sin^2 3\pi y) + (y - 1)^2 (1 + \sin^2 2\pi y)$ & (1,1) & $[-10,10]^2$ & $(1.000000, 1.000000)$ \\
\midrule
Easom & $-\cos(x) \cos(y) e^{-((x - \pi)^2 + (y - \pi)^2)}$ & $(\pi,\pi)$ & $[-100,100]^2$ & $(3.141594, 3.141593)$ \\
\bottomrule
\end{tabular*}
\caption{Comparison of test functions, their global minima, and the approximations found in a specific search space by UEPS.}
\label{tab:Ej_0}
\end{table}

\begin{table}[ht]
\centering
\scriptsize
\begin{tabular*}{\textwidth}{l @{\extracolsep{\fill}}llll}
\toprule
Function & Formula & Global minimum & Range & Approximation \\
\midrule
Ackley & $-20 e^{-0.2\sqrt{0.5(x^2 + y^2)}} - e^{0.5(\cos 2\pi x + \cos 2\pi y)} + e + 20$ & (0,0) & $[-5,5]^2$ & $(0.000000, 0.000000)$ \\
\midrule
Sphere & $x^2 + y^2$ & (0,0) & $[-100,100]^2$ & $(0.000000, 0.000000)$ \\
\midrule
Rosenbrock & $(1-x)^2+100(y-x^2)^2$ & (1,1) & $[-10,10]^2$ & $(1.000000, 1.000000)$ \\
\midrule
Beale & $(1.5 - x + xy)^2 + (2.25 - x + xy^2)^2 + (2.625 - x + xy^3)^2$ & (3,0.5) & $[-4.5,4.5]^2$ & $(3.500000, 0.614057)$ \\
\midrule
Booth & $(1.5 - x + xy)^2 + (2.25 - x + xy^2)^2$ & (1,3) & $[-10,10]^2$ & $(1.000000, 3.000000)$ \\
\midrule
Matyas & $0.26(x^2 + y^2) - 0.48xy$ & (0,0) & $[-10,10]^2$ & $(0.000000, 0.000000)$ \\
\midrule
Levy & $\sin^2 3\pi x + (x - 1)^2 (1 + \sin^2 3\pi y) + (y - 1)^2 (1 + \sin^2 2\pi y)$ & (1,1) & $[-10,10]^2$ & $(1.000000, 1.000000)$ \\
\midrule
Easom & $-\cos(x) \cos(y) e^{-((x - \pi)^2 + (y - \pi)^2)}$ & $(\pi,\pi)$ & $[-100,100]^2$ & $(3.000000, 3.171569)$ \\
\bottomrule
\end{tabular*}
\caption{Comparison of test functions, their global minima, and the approximations found in a specific search space by PSO.}
\label{tab:Ej_0_1}
\end{table}

\subsection{Problems with Constraints}
Next, we consider some optimization problems with constraints. Again, 10 runs of the algorithm have been performed using 50 particles and 100 iterations in each case.

\subsubsection{Rosenbrock Function with Constraints}
Consider the following problem: 

\begin{equation} 
\begin{aligned} \label{rosenbrock_restric}
&\ \min_{\mathbf{x} \in \mathbb{R}^n} f(x,y) = (1 - x)^2 + 100(y - x^2)^2\\
&\ \text{subject to:} \\
&\ -1.5 \leq x \leq 1.5,\\
&\ -0.5 \leq y \leq 2.5,\\
&\ (x - 1)^3 - y + 1 \leq 0,\\
&\ x + y - 2 \leq 0.
\end{aligned}
\end{equation}
The function \((1 - x)^2 + 100(y - x^2)^2\) is known as the \textit{Rosenbrock function}, and its global minimum is located at \((x^*, y^*) = (1, 1)\), with the value \(f(x^*, y^*) = 0\) (see the red point in figure \ref{fig:figura1}), which can be easily proven using elementary calculus tools. However, the minimum is located on a flat region shaped like a parabola, which presents a significant challenge for the convergence of deterministic methods.

\begin{figure}[htbp]
    \centering
    % Subfigura 1
    \begin{subfigure}[b]{0.45\textwidth} 
        \centering
        \includegraphics[width=\textwidth]{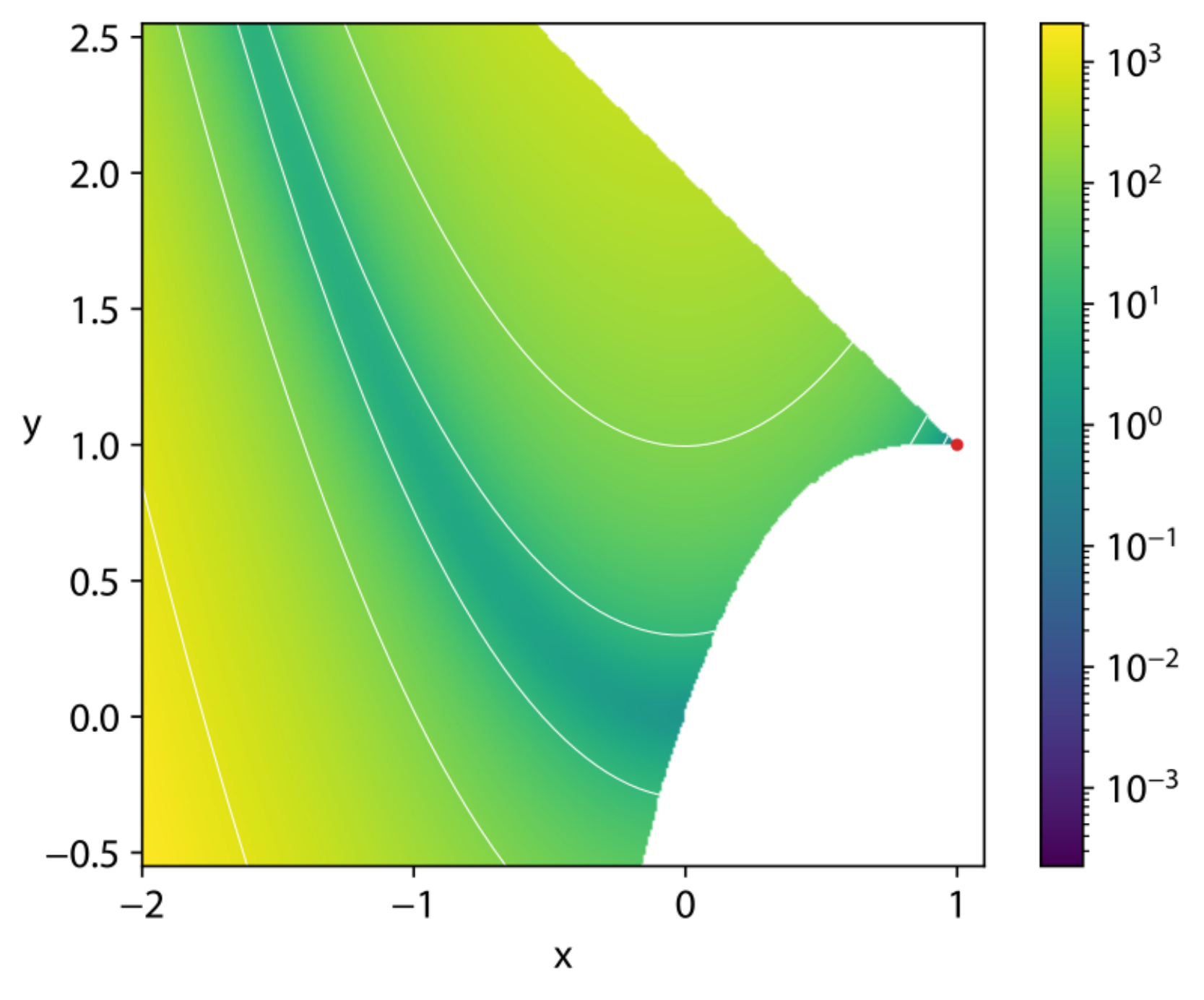} % Asegúrate de tener la imagen
        \caption{Location of the minimum in the search region.}
        \label{fig:figura1}
    \end{subfigure}
    \hspace{0.05\textwidth} % Espacio entre las figuras
    % Subfigura 2
    \begin{subfigure}[b]{0.4\textwidth} 
        \centering
        \includegraphics[width=\textwidth]{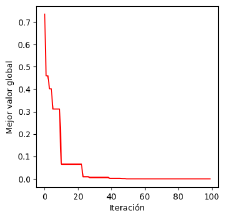} % Asegúrate de tener la imagen
        \caption{Algorithm performance.}
        \label{fig:figura2}
    \end{subfigure}
    \caption{Problema (\ref{rosenbrock_restric}).}
    \label{fig:figura_0}
\end{figure}

In order to address the problem posed in (\ref{rosenbrock_restric}), we have reformulated the function according to the strategy proposed in (\ref{eqOpt1}), assigning a unit weight to the constraints. Using 6 significant figures, 100 iterations, and only 50 particles, our metaheuristic provides the approximation \((1.000000, 1.000000)\). The algorithm's performance can be observed in figure~\ref{fig:figura2}.

\subsubsection{Pressure Vessel Design Problem}

The objective of this problem is to minimize costs, including material, forming, and welding expenses associated with a cylindrical vessel, as illustrated in figure \ref{fig:figura_0}. The vessel is closed at both ends with a hemispherical-shaped head. There are four key variables in this problem: $T_s$ ($x_1,$ shell thickness), $T_h$ ($x_2,$ head thickness), $R$ ($x_3,$ inner radius), and $L$ ($x_4,$ length of the cylindrical section of the vessel, excluding the head).

\begin{figure}[htbp]
    \centering
    % Subfigura 1
    \begin{subfigure}[b]{0.5\textwidth} 
        \centering
        \includegraphics[width=\textwidth]{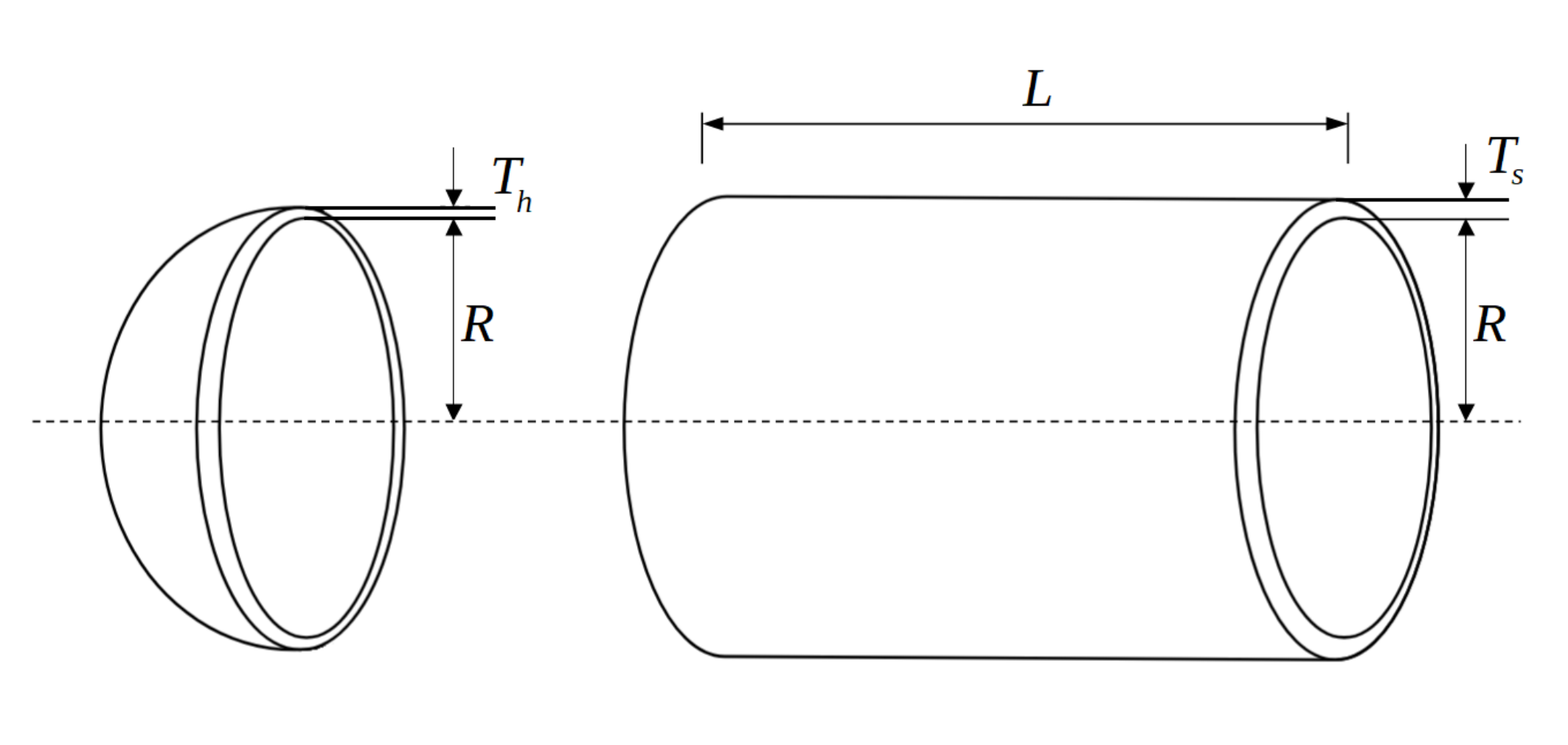} % Asegúrate de tener la imagen
        \caption{Pressure vessel design.}
        \label{fig:figura1}
    \end{subfigure}
    \hspace{0.05\textwidth} % Espacio entre las figuras
    % Subfigura 2
    \begin{subfigure}[b]{0.4\textwidth} 
        \centering
        \includegraphics[width=\textwidth]{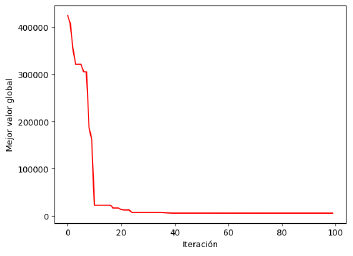} % Asegúrate de tener la imagen
        \caption{Algorithm performance.}
        \label{fig:figura2}
    \end{subfigure}
    \caption{Pressure vessel design problem.}
    \label{fig:figura_2}
\end{figure}

This problem is governed by four restrictions, and its formulation is articulated as follows:

\begin{align} \label{eq_1}
    & \min \quad  f(\mathbf{x}) = 0.6244x_1x_3x_4 + 1.7781x_2x_3^2 +3.1661x_1^2x_4 + 19.84x_1^2x_3  \\
    & \text{Sujeto a:}  \nonumber \\ 
    & g_1(\mathbf{x}): -x_1 + 0.0193x_3\leq 0 \nonumber \\
    & g_2(\mathbf{x}): -x_2 + 0.00954x_3\leq 0 \nonumber \\
    & g_3(\mathbf{x}): -\pi x_3^2x_4 - \frac{4}{3}\pi x_3^3 +1296000 \leq 0 \nonumber \\ 
    & g_4(\mathbf{x}): x_4 - 240\leq 0 \nonumber \\
    & 0 \leq x_1 \leq 99 \nonumber \\
    & 0 \leq x_2 \leq 99 \nonumber \\
    & 10 \leq x_3 \leq 200 \nonumber \\
    & 10 \leq x_4 \leq 200. \nonumber 
\end{align}

In order to solve problem (\ref{eq_1}), we transform it into an unrestricted problem following strategy (\ref{eqOpt2}), with four inequality constraints. 

\begin{table}[ht]
{\small
    \noindent\begin{tabular*}{\textwidth}{l @{\extracolsep{\fill}}lllll}
    \toprule
    \multicolumn{1}{l}{\small Algorithm}& \multicolumn{4}{l}{\small Optimal variables}& \multicolumn{1}{l}{\small Optimal costo}\\
    \cline{2-5} % Línea divisoria desde la columna 2 hasta la 6
        &$T_s$& $T_h$& $R$& $L$& \\
      \textbf{UEPS}& \textbf{0.778169}& \textbf{0.384698}& \textbf{40.319619}& \textbf{200.000000}& \textbf{5885.473070} \\
      AEO& 0.8374205& 0.413937& 43.389597& 161.268592& 5994.50695 \\   
      BA& 0.812500& 0.437500& 42.098445& 176.636595& 6059.7143 \\
      CSS& 0.812500& 0.437500& 42.103624& 176.572656& 6059.0888 \\
      FHO& 0.8375030& 0.4139782& 43.3939372& 161.2185336& 5994.6845509 \\
      GA& 0.812500& 0.437500& 42.097398& 176.654050& 6059.9463 \\
      GPEAae& 0.812500& 0.437500& 42.098497& 176.635954& 6059.708025 \\ 
      G-QPSO& 0.812500& 0.437500& 42.0984& 176.6372& 6059.7208 \\
      GWO& 0.8125& 0.4345& 42.089181& 176.758731& 6051.5639 \\
      HHO& 0.81758383& 0.4072927& 42.09174576& 176.7196352& 6000.46259 \\
      HPO& 0.778168& 0.384649& 40.3196187& 200& 5885.33277 \\
      HPSO& 0.812500& 0.437500& 42.0984& 176.6366& 6059.7143 \\
      MFO& 0.8125& 0.4375& 42.098445& 176.636596& 6059.7143 \\
      SC-GWO& 0.8125& 0.4375& 42.0984& 176.63706& 6059.7179 \\
      WEO& 0.812500& 0.437500& 42.098444& 176.636622& 6059.71 \\
      WOA& 0.812500& 0.437500& 42.0982699& 176.638998& 6059.7410 \\
     \hline
    \end{tabular*}
    }
    \caption{Comparison of numerical results of different bio-inspired algorithms on the pressure vessel design problem.}
    \label{tab:Ej_1}   
\end{table}

The comparative results with other algorithms are presented in table \ref{tab:Ej_1}. The evaluated algorithms include the \textit{Fuzzy Hunter Optimizer (FHO)}, \textit{Artificial Ecosystem Optimization (AEO)}, \textit{Bat Algorithm (BA)}, \textit{Charged System Search (CSS)}, \textit{Genetic Algorithm (GA)}, \textit{Gray Prediction Evolutionary Algorithm Based on Uniform Acceleration (GPEAae)}, \textit{Quantum-Gaussian Particle Swarm Optimization (G-QPSO)}, \textit{Gray Wolf Optimization Algorithm (GWO)}, \textit{Harris Hawks Optimizer (HHO)}, \textit{Hunter Optimizer (HPO)}, \textit{Hybrid Particle Swarm Optimization (HPSO)}, \textit{Moth-Flame Optimization (MFO)}, \textit{Gray Wolf Optimizer with Sine-Cosine (SC-GWO)}, \textit{Water Evaporation Optimization (WEO)}, and \textit{Whale Optimization Algorithm (WOA)}. In all cases, we used a set of 50 particles and 100 iterations in our algorithm, while the compared algorithms employed more than 100 individuals in their respective executions. Despite this difference in the number of particles, the performance of our algorithm was highly competitive, being surpassed only by the \textit{FHO}. The execution time of the algorithm was 0.285184 seconds. Finally, figure \ref{fig:figura_2} illustrates the performance of our algorithm throughout the iterations. Increasing the number of particles and/or iterations can yield better results.

\section{Conclusion and Future Work}

In this paper, we have presented a modified version of PSO, inspired by underdamped vibrations. This modification allows particles to dynamically explore the search space, alternating between exploration and exploitation processes. Furthermore, the ability of particles to temporarily exceed the optimal solution facilitates the exploration of new regions, preventing them from getting trapped in local optima.

The numerical experiments conducted have demonstrated that our metaheuristic is highly efficient, both in constrained and unconstrained optimization problems. In fact, in the pressure vessel design problem, the performance of our approach surpassed that of many well-known and widely established metaheuristics.

As future work, we aim to validate our metaheuristic on a broader range of optimization problems, in order to assess its performance in diverse contexts. Additionally, we will conduct a more thorough statistical study to better understand the advantages and limitations of our approach in comparison with other optimization techniques.

% Apéndice
\appendix

\begin{appendices}

\section{Python implementation}

In this section, the process of implementing the UEPS algorithm will be detailed. As previously mentioned, two main modules are used. The first of these modules is \texttt{test\_functions.py}, in which the test functions and classic optimization problems used to validate the proposed metaheuristic in this paper are defined. These problems include both unconstrained functions and problems with constraints, each associated with a specific search domain.

The second module, \texttt{oeps.py}, implements the UEPS metaheuristic. This module imports the first one to access the test functions and definitions of the optimization problems. Through the variable \texttt{func\_name}, the user can specify the optimization function or problem they wish to address. In this way, the UEPS algorithm can be easily adapted to different contexts without the need to modify its internal code.

The workflow of the algorithm begins with the initialization of the particles in a search space determined by the lower ($L_b$) and upper ($U_b$) bounds of the selected function variables. Then, the algorithm iteratively optimizes the position of the particles through a combination of oscillations and random perturbations, adjusting their velocity and position at each iteration. The best value found during this process is reported at the end of the execution.

\subsection{Library Installation}

To start working with the project, it is necessary to install the required libraries, either in a virtual environment or globally. In a distribution such as Debian 12, for example, this can be easily done from the terminal with the following command:

\begin{lstlisting} 
pip install numpy plotext tqdm 
\end{lstlisting}

This command will install the \texttt{numpy}, \texttt{plotext}, and \texttt{tqdm} libraries, which are essential for the operation of the project. If these libraries are not already installed, the command will automatically download and install them.

\subsection{Code of \texttt{test\_functions.py}}

\begin{lstlisting}[language=Python]
    import numpy as np

# Funciones de testeo sin restricciones
def ackley (x):
    f_1 = -20 * np.exp(-0.2 * np.sqrt(0.5 * (x[0]**2 + x[1]**2)))
    f_2 = -np.exp(0.5 * (np.cos(2 * np.pi * x[0]) + np.cos(2 * np.pi * x[1]))) + np.exp(1) + 20
    return f_1 + f_2

def sphere(x):
    return np.sum(x**2)

def rosenbrock(x):
    n = len(x)
    f = 0
    for i in range(n - 1):
        f += 100 * (x[i + 1] - x[i] ** 2) ** 2 + (1 - x[i]) ** 2
    return f

def beale(x):
    f_1 = (1.5 - x[0] + x[0] * x[1])**2
    f_2 = (2.25 - x[0] + x[0] * x[1]**2)**2
    f_3 = (2.625 - x[0] + x[0] * x[1]**3)**2
    return f_1 + f_2 + f_3

def booth(x):
    return (x[0] + 2 * x[1] - 7)**2 + (2 * x[0] + x[1] - 5)**2

def bukin_function_N6(x):
    return 100 * np.sqrt(np.abs(x[1] - 0.01 * x[0]**2)) + 0.01 * np.abs(x[0] + 10)

def matyas(x):
    return 0.26 * (x[0]**2 + x[1]**2) - 0.48 * x[0] * x[1]

def levy(x):
    f_1 = np.sin(3 * np.pi * x[0])**2
    f_2 = (x[0] - 1)**2 * (1 + np.sin(3 * np.pi * x[1])**2)
    f_3 = (x[1] - 1)**2 * (1 + np.sin(3 * np.pi * x[1])**2)
    return f_1 + f_2 + f_3

def easom(x):
    return -np.cos(x[0]) * np.cos(x[1]) * np.exp(-((x[0] - np.pi)**2 + (x[1] - np.pi)**2))

def eggholder(x):
    f_1 = -(x[1] + 47) * np.sin(np.sqrt(np.abs(x[0] / 2 + (x[1] + 47))))
    f_2 = -x[0] * np.sin(np.abs(x[0] - (x[1] + 47)))
    return f_1 + f_2

def mccormick(x):
    return np.sin(x[0] + x[1]) + (x[0] - x[1])**2 - 1.5 * x[0] + 2.5 * x[1] + 1

def eggcrate(x):
    return x[0]**2 + x[1]**2 + 25 * (np.sin(x[0])**2 + np.sin(x[1])**2)

def michalewicz(x):
    n = len(x)
    f = 0
    for i in range(n):
        f += np.sin(x[i]) * (np.sin(((i + 1) * x[i]**2) / np.pi))**20
    return -f

# Funciones de testeo con restricciones
def rosenbrock_constrained(x):
    n = len(x)
    f = 0
    for i in range(n - 1):
        f += 100 * (x[i + 1] - x[i] ** 2) ** 2 + (1 - x[i]) ** 2
        f += max((x[0] - 1)**3 - x[1] + 1, 0) + max(x[0] + x[1] - 2, 0)
    return f

# Pressure vessel design problem
def pressure_vessel_design_problem(x):
    f = 0.6224 *x[0] * x[2] * x[3] + 1.7781 * x[1] * (x[2]**2) \
        + 3.1661 * (x[0]**2) * x[3] + 19.84 * (x[0]**2) * x[2]
    g1 = max(-x[0] + 0.0193 * x[2], 0)
    g2 = max(-x[1] + 0.00954 * x[2], 0)
    g3 = max(-np.pi * (x[2]**2) * x[3] - (4 / 3) * np.pi * (x[2]**3) + 1296000, 0)
    g4 = max(x[3] - 240, 0)
    #------------------- Método de penalidad estática----------------------
    K = 10**9
    if g1 + g2 + g3 + g4 == 0:
        fitness = f
    else:
        suma = 0
        if g1 != 0:
            suma = suma + 1
        if g2 != 0:
            suma = suma + 1
        if g3 != 0:
            suma = suma + 1
        if g4 != 0:
            suma = suma + 1
        fitness = K - suma * K / 4
    f = fitness
    return f

# Tension/compression spring problem
def tension_compression_spring_problem(x):
    f = (x[2] + 2) * x[1] * x[0]**2
    # Restricciones
    g1 = max(1 - (x[2] * x[1]**3) / (71785 * x[0]**4), 0)

    g2 = (4 * x[1]**2 - x[0] * x[1]) / (12566 * (x[1] * (x[0]**3) - x[0] **4))
    g2 = g2 + 1 / (5108 * (x[0]**2)) - 1
    g2 = max(g2, 0)

    g3 = 1 - (140.45 * x[0]) / (x[2] * x[1]**2)
    g3 = max(g3, 0)

    g4 = (x[0] + x[1]) / 1.5 - 1
    g4 = max(g4, 0)

    G = [g1, g2, g3, g4]

    #------------------- Método de penalidad estática----------------------
    K = 10**9
    if sum(G) == 0:
        fitness = f
    else:
        suma  = 0
        if g1 != 0:
            suma = suma + 1
        if g2 != 0:
            suma = suma + 1
        if g3 != 0:
            suma = suma + 1
        if g4 != 0:
            suma = suma + 1
        fitness = K - suma * K / 4
    f = fitness
    return f

# Estructura que contiene funciones y sus regiones de búsqueda
functions_with_bounds = {
    'ackley': (ackley, np.array([-5, -5]), np.array([5, 5])),
    'sphere': (sphere, np.array([-100, -100]), np.array([100, 100])),
    'rosenbrock': (rosenbrock, np.array([-10, -10]), np.array([10, 10])),
    'beale': (beale, np.array([-4.5, -4.5]), np.array([4.5, 4.5])),
    'booth': (booth, np.array([-10, -10]), np.array([10, 10])),
    'bukin_function_N6': (bukin_function_N6, np.array([-15, -3]), np.array([-5, 3])),
    'matyas': (matyas, np.array([-10, -10]), np.array([10, 10])),
    'levy': (levy, np.array([-10, -10]), np.array([10, 10])),
    'easom': (easom, np.array([-100, -100]), np.array([100, 100])),
    'eggholder': (eggholder, np.array([-512, -512]), np.array([512, 512])),
    'mccormick': (mccormick, np.array([-1.5, -3]), np.array([4, 4])),
    'eggcrate': (eggcrate, np.array([-5, -5]), np.array([5, 5])),
    'michalewicz': (michalewicz, np.array([0, 0]), np.array([np.pi, np.pi])),
    'rosenbrock_constrained': (rosenbrock_constrained, np.array([-1.5, -0.5]), np.array([1.5, 2.5])),
    'pressure_vessel_design_problem': (pressure_vessel_design_problem, np.array([0, 0, 10, 10]), np.array([99, 99, 200, 200])),
    'tension_compression_spring_problem': (tension_compression_spring_problem, np.array([0.05, 0.25, 2]), np.array([2, 1.30, 15])),
}
\end{lstlisting}

\subsection{Code of \texttt{oeps.py}}

\begin{lstlisting}[language=Python]

import numpy as np
import plotext as plt
from tqdm import tqdm
import time
from test_functions import functions_with_bounds

# Iniciar el temporizador
start_time = time.time()

# Presentación del programa
print("=======================================================")
print("Optimización por Enjambre de Partículas Subamortiguadas")
print("=======================================================")
print()
print("Autor: Dr. Hernández Rodríguez, Matías Ezequiel")
print("Correo: matiasehernandez@gmail.com")
print("Copyright @ 2024 El Túnel del Misterio. Todos los derechos reservados.")
print("Uso: Este material está destinado exclusivamente a fines educativos y no comerciales")
print("-------------------------------------------------------------------------------------")
print()

# Nombre de la función objetivo
func_name = 'pressure_vessel_design_problem' # Una de las funciones de testeo

# Acceder a la función y cotas del espacio de búsqueda
objective_function, L_b, U_b = functions_with_bounds[func_name]

# Parámetros del algoritmo
A = 1 # Amplitud de oscilación
b = 0.007 # Tasa de amortiguamiento
w_min = 0.4 # Factor de inercia mínimo
w_max = 0.9 # Factor de inercia máximo
alpha = 0.8 # Factor de perturbación aleatoria
dim = len(L_b) # Dimensión del espacio de búsqueda (puedes cambiarlo para más dimensiones)
max_iter = 100 # Número máximo de iteraciones
n_particles = 50 # Número de partículas

# Inicialización de partículas y velocidad
np.random.seed(42) # Fijar semilla para reproducibilidad
x_i = np.random.uniform(L_b, U_b, (n_particles, dim)) # Posiciones iniciales aleatorias
v_i = np.zeros_like(x_i) # Velocidades iniciales (cero)
x_best = x_i.copy() # Mejor posición personal
f_best = np.array([objective_function(x) for x in x_best]) # Mejor valor de la función objetivo
x_global_best = x_best[np.argmin(f_best)] # Mejor posición global
f_global_best = np.min(f_best) # Mejor valor de la función objetivo global

# Función de actualización de las partículas
def update_position(x_i, v_i, x_global_best, t):
    r = np.random.rand(x_i.shape[0]) # Vector aleatorio entre 0 y 1 para oscilaciones
    oscillation_term = A * (1 - np.cos(2 * np.pi * r)) * np.exp(-b * t) # Término oscilatorio

    rand_term = alpha**t * (np.random.rand(x_i.shape[0]) - 0.5) # Término aleatorio adicional
    new_v = (w_max - (w_max - w_min) * t / max_iter) * v_i + oscillation_term[:, None] * (x_global_best - x_i) + rand_term[:, None] # Actualizar velocidad
    new_x = x_i + new_v # Actualizar posición

    # Limitar la posición a los límites permitidos
    new_x = np.clip(new_x, L_b, U_b)

    return new_x, new_v

# Para almacenar el progreso de la mejor función objetivo a lo largo de las iteraciones
progress = []

# Ciclo de optimización
for t in tqdm(range(max_iter), desc='Progress', dynamic_ncols=True):
    # Actualizar posiciones y velocidades
    x_i, v_i = update_position(x_i, v_i, x_global_best, t)

    # Evaluar la función objetivo en las nuevas posiciones
    f_new = np.array([objective_function(x) for x in x_i])

    # Actualizar el mejor conocido para cada partícula
    mask = f_new < f_best
    x_best[mask] = x_i[mask]
    f_best[mask] = f_new[mask]

    # Actualizar la mejor solución global
    min_f_best = np.min(f_best)
    if min_f_best < f_global_best:
        f_global_best = min_f_best
        x_global_best = x_best[np.argmin(f_best)]

    # Almacenar el progreso
    progress.append(f_global_best)

# Resumen del problema y resultados numéricos
print()
print("========================================")
print("PARÁMETROS USADOS Y RESULTADOS NUMÉRICOS")
print("========================================")
print()

# Función objetivo
print("Objective function")
print("------------------")
# Obtener el nombre de la función
func_name = func_name
# Reemplazar la primera letra por la misma en mayúscula
modified_name = func_name[0].upper() + func_name[1:]
print(f"f(x): {modified_name}")
print()

# Parámetros usados
print("Parámetros usados")
print("-----------------")
print(f"Amplitud de oscilación: {A}")
print(f"Tasa de amortiguamiento: {b}")
print(f"Factor de perturbación aleatoria: {alpha}")
print(f"Factor de inercia mínimo: {w_min}")
print(f"Factor de inercia máximo: {w_max}")
print(f"Número de partículas: {n_particles}")
print(f"Número de iteraciones: {max_iter}")
print(f"Número de variables: {dim}")
print(f"Límite inferior para las variables: {L_b}")
print(f"Límite superior para las variables: {U_b}")
print()

print("Resultados Numéricos")
print("--------------------")
print(f"x*: ({', '.join([f'{x:.6f}' for x in x_global_best])})")
print(f"f(*): {f_global_best:.6f}")
print()

# Graficar vector de costos
print("Rendimiento del algoritmo")
print("-------------------------")

# Visualización del progreso
plt.clear_figure()
plt.plotsize(60, 20)  # Ancho y alto en caracteres
plt.plot(progress, color='red')
plt.xlabel("Iteración")
plt.ylabel("Mejor valor global")
plt.title("Progreso de la optimización")
plt.theme("matrix") # Otras opciones: pro, dark
plt.show()

print("Tiempo de ejecución de PSOSA")
print("----------------------------")
# Calcular el tiempo transcurrido
end_time = time.time()

# Imprimir el tiempo de ejecución
print(f"El algoritmo se ejecutó en: {end_time - start_time} segundos")
\end{lstlisting}

\end{appendices}

% Bibliografía

\end{document}